\pdfoutput=1

\documentclass[letterpaper, 10 pt, conference]{ieeeconf}  

\IEEEoverridecommandlockouts                              

\usepackage{longtable,tabularx}
\usepackage{tcolorbox}
\usepackage{xcolor,colortbl}
\usepackage{diagbox}
\usepackage{multirow}

\usepackage{graphicx}
\usepackage{subcaption}
\usepackage{epsfig} 
\usepackage{amsmath} 
\usepackage{amssymb}  
\usepackage{color}
\usepackage{cite}
\usepackage[ruled, linesnumbered]{algorithm2e}

\usepackage{amsthm}
\usepackage[pscoord]{eso-pic} 
\newtheorem{definition}{Definition}
\newtheorem{problem}{Problem}
\newtheorem{example}{Example}

\newcommand{\placetextbox}[3]{
  \setbox0=\hbox{#3}
  \AddToShipoutPictureFG*{
    \put(\LenToUnit{#1\paperwidth},\LenToUnit{#2\paperheight}){\vtop{{\null}\makebox[0pt][c]{#3}}}%
  }%
}%

\DeclareMathOperator{\Val}{Val}
\DeclareMathOperator{\reg}{reg}
\newcommand{\lang}{\mathcal{L}}

\newcommand{\B}{\mathcal{B}}
\newcommand{\G}{\mathcal{G}}
\newcommand{\PA}{\mathcal{P}}
\newcommand{\Tau}{\mathrm{T}}
\renewcommand{\phi}{\varphi}

\definecolor{Gray}{gray}{0.85}
\definecolor{LightCyan}{rgb}{0.88,1,1}

\newcolumntype{a}{>{\columncolor{Gray}}c}
\newcolumntype{b}{>{\columncolor{white}}c}

\title{\LARGE \bf
Let's Collaborate: Regret-based Reactive Synthesis for \\ Robotic Manipulation
}

\author{Karan Muvvala, Peter Amorese, and Morteza Lahijanian
\thanks{This work was supported in part by University of Colorado Boulder and NASA COLDTech Program under grant \#80NSSC21K1031.}%
\thanks{Authors are with the department of Aerospace Engineering Sciences at the University of Colorado Boulder, CO, USA
        {\tt\small \{\textit{firstname}.\textit{lastname}\}@colorado.edu}}%
}

\begin{document}
\placetextbox{0.5}{0.95}{To appear in IEEE International Conference on Robotics and Automation (ICRA), May. 2022.}
\maketitle



\begin{abstract}

As robots gain capabilities to enter our human-centric world, they require formalism and algorithms that enable smart and efficient interactions. This is challenging, especially for robotic manipulators with complex tasks that may require collaboration with humans. Prior works approach this problem through reactive synthesis and generate strategies for the robot that guarantee task completion by assuming an adversarial human. While this assumption gives a sound solution, it leads to an ``unfriendly'' robot that is agnostic to the human intentions. We relax this assumption by formulating the problem using the notion of \emph{regret}.  We identify an appropriate definition for regret and develop regret-minimizing synthesis framework that enables the robot to seek cooperation when possible while preserving task completion guarantees. We illustrate the efficacy of our framework via various case studies.



\end{abstract}


\section{Introduction}
\label{sec: intro}



From factories to households, robots are rapidly leaving behind their robot-centric environments and entering our society.
Examples include self-driving cars, delivery robots, and assistive robots.
To be successful in our human-centric world, robots must develop the ability to interact with dynamic environments. This includes performing complex tasks in the presence of humans, who have their own objectives and may interfere with the robot's task.
Therefore, robots must aim to have effective interactions with humans and seek collaboration whenever possible.
To achieve such capabilities, strategies that account for human objectives as well as task and resource constraints are needed. 
Generating such strategies, however, is challenging due to two main reasons: \emph{formulation} and \emph{computation}. That is, a proper mathematical formulation of such strategies is a nontrivial problem, and computation cost for strategies that enable reactivity is inherently high, especially in the manipulation domain, where tasks are complex and space of reasoning is high dimensional. This work mainly focuses on the formulation challenge and also aims to design a reactive synthesis framework that enables robots to seek collaboration with humans while guaranteeing completion of their task.



\begin{figure}[t]
    \begin{subfigure}[t]{0.5\linewidth}
        \includegraphics[width=0.98\linewidth]{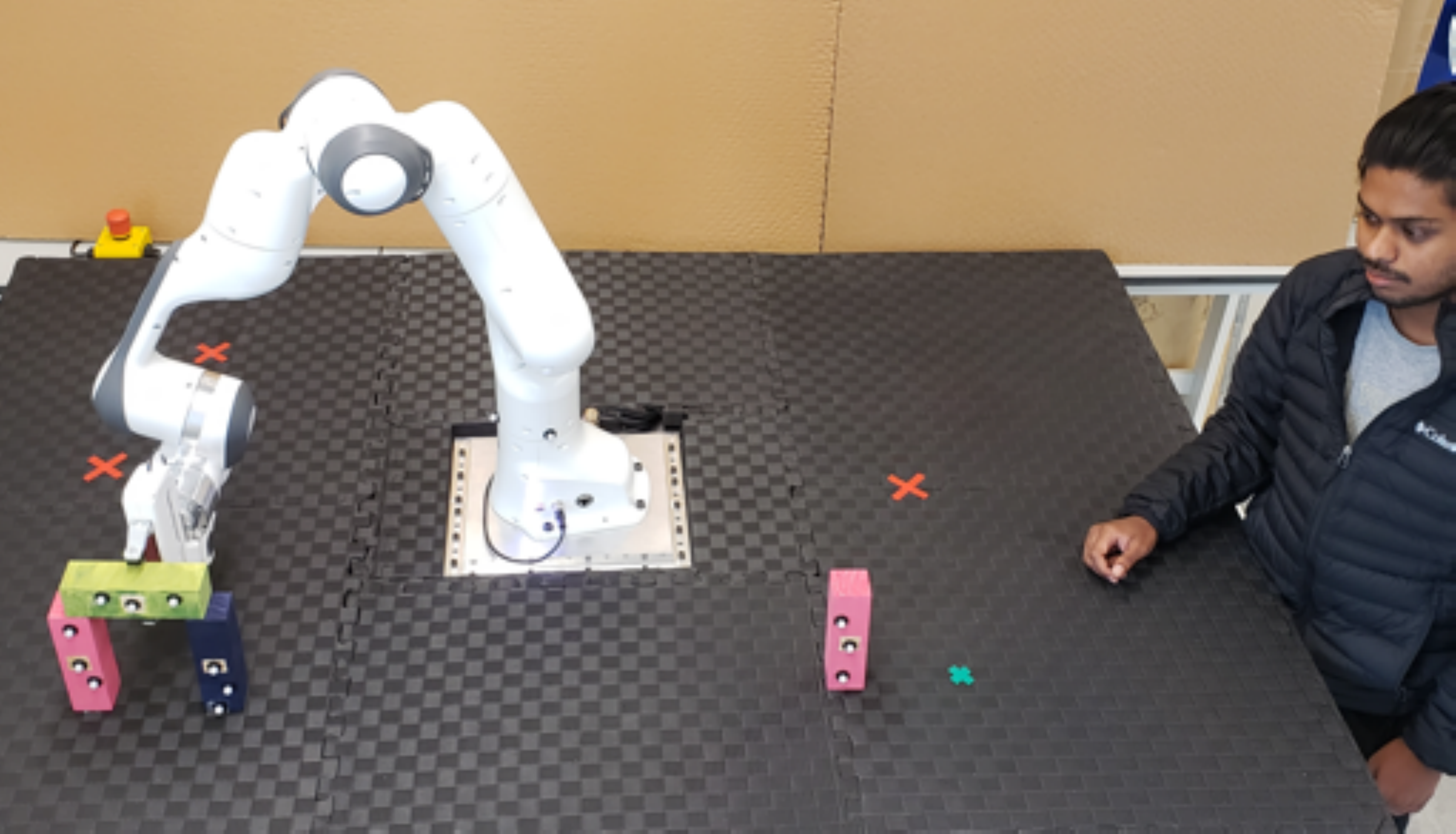}
        \caption{Adversarial Behavior}
        \label{fig: adv_beh}
    \end{subfigure}%
    \begin{subfigure}[t]{0.5\linewidth}
        \centering
        \includegraphics[width=0.98\linewidth]{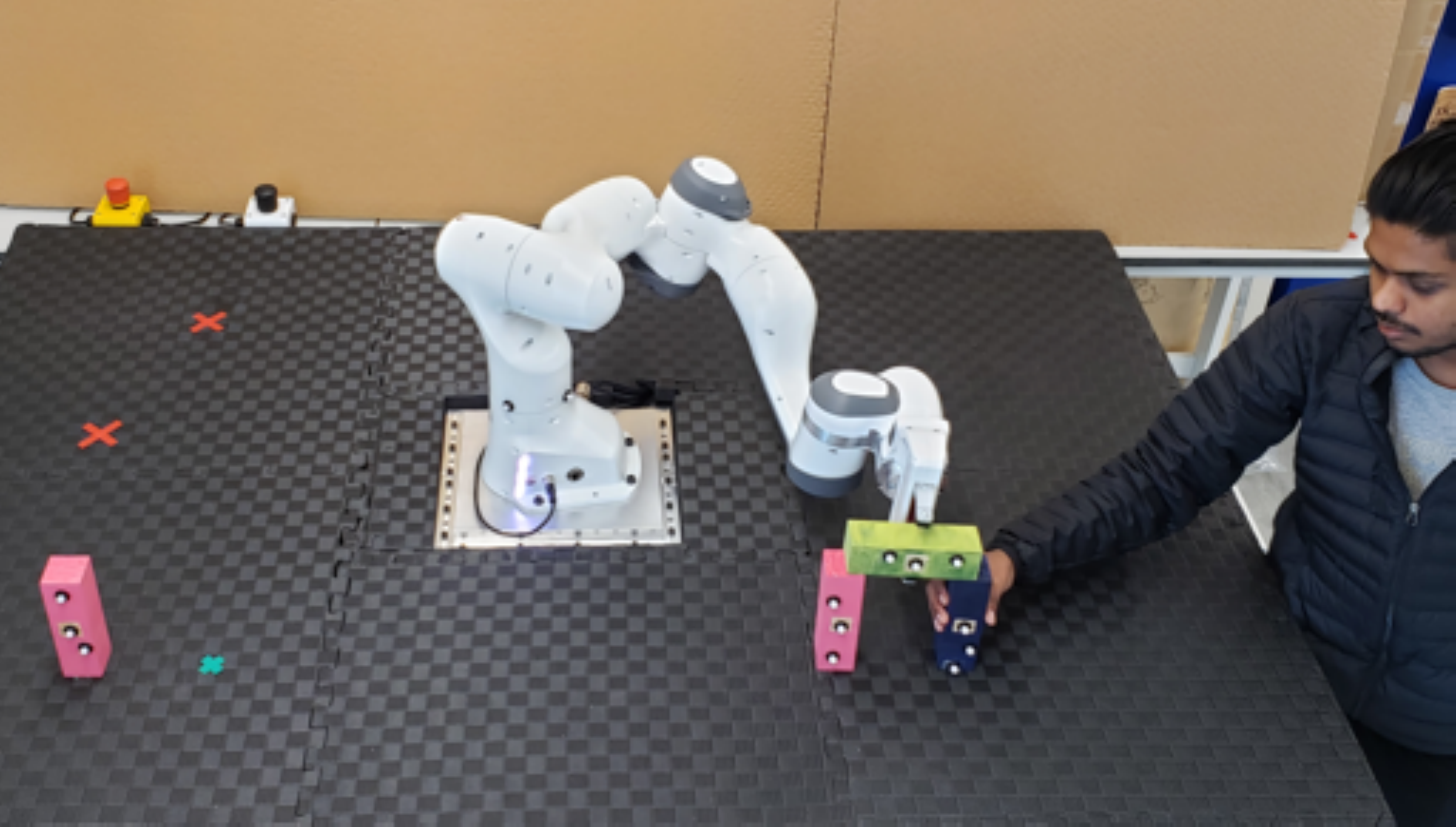}
        \caption{Probabilistic Behavior}
        \label{fig: prob_beh}
    \end{subfigure}%
    \caption{Arch Construction. 
    (a) Strategy via \cite{he2017reactive} (human assumed to be adversarial).
    (b) Policy via \cite{wells2021probabilistic} (robot expects human to be cooperative based on prior data).
    }
    \label{fig: reg_illustration}
    \vspace{-5mm}
\end{figure}

As an example, consider the scenario in Fig. \ref{fig: reg_illustration}, 
where the robot is tasked with building an arch either on the left or right side of the table with a green block on top.
In this workspace, a human can reach and manipulate the blocks placed on the right side but not the ones on the left. To operate in the left side, the robot has to spend more energy than the one closer to the human. For such scenarios, classical planning methods that compute a fixed sequence of actions are not sufficient.
Instead, we need the robot to reason and react to the human's actions for efficient 
interactions.  


Previous work \cite{he2015towards, he2017reactive, he2019automated} addresses this problem through the lens of \textit{reactive synthesis} \cite{hadas2012reactive, belta2014reactive, wolff2013reactive}. They employ \emph{Linear Temporal Logic over finite traces} (LTLf) \cite{vardi2013ltlf} to specify tasks that can be accomplished in finite time. The framework models the interaction between the human and the robot as a two-player game. 
By assuming the human to be purely adversarial in this game, a strategy is synthesized for the robot that guarantees task completion under all possible human moves. This assumption however is conservative,  eliminates any room for collaboration, and results in ``unfriendly'' behaviors that could lead to higher energy spending than allowing a chance for collaboration.  
In Fig. \ref{fig: adv_beh}, using that approach, the robot builds the arch away from the human irrespective of the human's intention. 


Recent work \cite{wells2021probabilistic} relaxes this assumption by modelling the human as a probabilistic agent. This leads to an abstraction in the form of a Markov Decision Process (MDP), and the objective reduces to synthesizing an optimal policy that maximizes the probability of satisfying the task. 
The approach optimizes the robot's actions according to the expected behavior of the human instead of assuming they are always adversarial. In the scenario in Fig.~\ref{fig: reg_illustration}, 
using this approach,
the robot builds the arch near the human  (Fig. \ref{fig: prob_beh}) if the expected behavior of the human is to be cooperative; otherwise, the robot builds the arch away from the human. While efficient interactions can be achieved using that framework, the required prior knowledge on the human is generally hard to obtain.  
The method also fails to capture the human as a strategic agent with their own objectives.


In the machine learning and game theoretic communities, an emerging method of reasoning about the quality of strategies (actions) is via the notion of regret \cite{chen2021minimax, levine2018regret, zinkevich2007regret, azar2017minimax,filiot2010iterated}. 
\textit{Regret} is the measure of ``goodness'' of an action in comparison to the best response that could have been  received in hindsight \cite{chen2021minimax, chen2021finding}.  This is different from the classical notion of cost or reward. 
In regret games, instead of trying to minimize the total cost, the objective is to minimize regret. 
The obtained strategies have shown to be more reasonable than, e.g., Nash Equilibrium strategies, for games played for finite number of cycles \cite{halpern2012iterated}.  
Various formalisms for regret have been introduced in different communities.
The reinforcement learning community specifically uses a formalism that is suitable for exploitation of the degree of incomplete information in games in regards to partial observability or uncertainty in transition probabilities \cite{jin2018regret, zinkevich2007regret}. 
Those approaches look at regret locally, whereas  
the formal methods community views regret globally through the lens of reactive systems \cite{guillermo2017reactive, guillermo2015discounted}.  
Nevertheless, the notion of regret has not been studied for robotic manipulation, especially in the context of reactive synthesis for tasks with human interventions.

In this work, we propose a regret-based reactive synthesis framework that enables a robotic manipulator to seek collaboration with the human while guaranteeing task completion and staying within its resource limits.  This framework relaxes the assumption that the human is purely adversarial while still capturing the human as a strategic agent without requiring \emph{a priori} knowledge. As in \cite{he2015towards, he2017reactive, he2019automated, wells2021probabilistic}, we consider tasks given as LTLf formulas.  Our approach is based on abstracting the interaction between the robot and human as a two-player game. 
We then formulate a regret game on the abstraction by defining an appropriate formalism for regret in the context of interactive manipulation domain. 
We adapt an algorithmic approach to generate regret minimizing strategies if they exist.  
We show that these strategies guarantee task completion and enable efficient interactions, but they are history dependent, which means they are computationally expensive.
Finally, we illustrate the benefits of regret-minimizing strategies in several case studies and compare the results against adversarial strategies.

The contributions of this work are threefold. First, we identify an appropriate regret formalism for the manipulation domain, and based on that,  we introduce a regret-minimizing reactive synthesis framework that encourages collaboration while still guaranteeing the satisfaction of the task and resource requirements.  This framework paves the way for further studies on using regret in robotics since regret-minimizing behaviors are natural and human-like \cite{halpern2012iterated}.  Second, we provide an end-to-end toolbox for synthesizing regret minimizing strategies.  Finally, we illustrate the regret-minimizing strategies and their corresponding emergent behaviors through several case studies \cite{videos, repolink}.
\section{Problem Formulation}
\label{sec: problem_formulation}


The goal of this work is to synthesize a high-level strategy for a robotic manipulator to achieve a task defined over a set of objects via an efficient interaction with a human. We assume that the human has their own objective that may not be necessarily adversarial to the robot's task. Hence, we want to enable the robot to explore possible collaboration with the human in the execution of the task while guaranteeing task completion.  
Below, we formalize this problem by first introducing the required mathematical definitions.



\begin{figure}[t]
    \centering
    \includegraphics[scale=0.27]{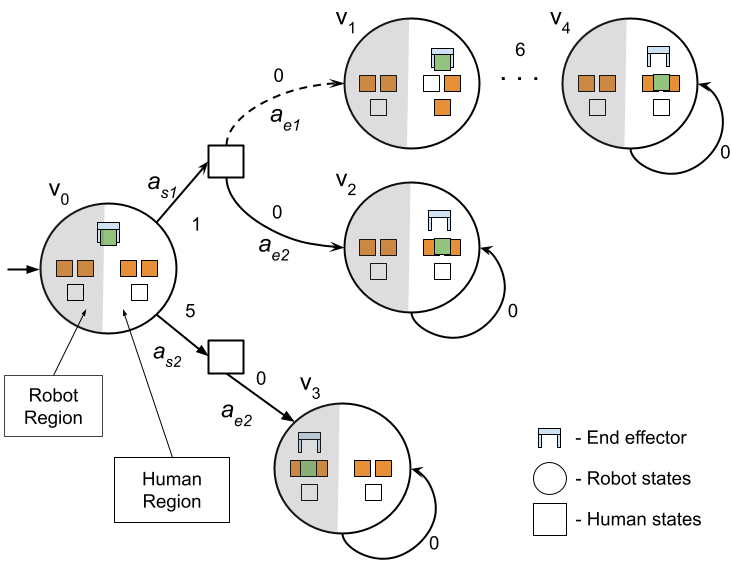}
    \caption{Example illustration of abstraction $\G$ for the Arch construction example in Fig.~\ref{fig: reg_illustration}. The human can move objects (colored blocks) in its region (unshaded) but cannot reach the objects placed in robot (shaded) region. Robot actions  $a_{s1}$ and $a_{s2}$ denote placing the green block in the human region and robot region, respectively. The edge weights are the transition (action) costs. 
    }
    \label{fig: Abstraction illustration}
    \vspace{-5mm}
\end{figure}

\subsection{Manipulation Domain Abstraction}
\label{ssec: manipulation_domain}

The manipulation domain describes how the robot, human, and objects can interact with each other. This domain is naturally continuous, and previous works \cite{he2017reactive, he2019automated} show how a discrete abstraction of it can be constructed.  The abstraction captures the interaction as a turn-based two-player game between the robot and human at discrete time steps.  

\begin{definition}[Manipulation Domain Abstraction] The manipulation domain abstraction is a tuple $\mathcal{G} = (V, v_0, A_s, A_e, \delta, F, \Pi, L)$ where:

\begin{itemize}
    \item $V = V_s \cup V_e$ is the set of states partitioned into robot (system) states $V_s$ and human (environment) states $V_e$,
    \item $A_s$ and $A_e$ are the sets of finite actions for the robot and human, respectively,
    \item $\delta: V \times (A_s \cup A_e) \to V$ is the transition function,
    \item $F: V\times (A_s \cup A_e) \times V \to \mathbb{R}_{\geq 0}$ is the cost function that maps each transition to a cost.  Here, the cost of transitions enabled by the human actions are assumed to be zero, i.e., $F(v,a_e,v') = 0$ $\;\forall a_e \in A_e$
    , 
    \item  $\Pi$ is a set of task related atomic propositions, and
    \item $L: V \to 2^\Pi$ is a labeling function that indicates the property of each state relative to the task.
\end{itemize}
\label{def: two_player_game}
\end{definition}

Each state of $\G$ represents
a configuration of the world, i.e., object locations and robot end-effector, and each transition corresponds to an action taken by the robot agent or the human agent. A transition from one state to another represents evolution of the shared workspace under that action. 


\begin{example}
    Fig. \ref{fig: Abstraction illustration} shows a simple abstraction $\G$ for the scenario in Fig.~\ref{fig: reg_illustration}. The robot starts from state $v_0$ with the green block in its end-effector. The robot has two choices from $v_0$: place the block near the human (unshaded region) or away from the human (shaded region). 
    Once the robot takes an action, it is the human's turn to decide whether to intervene or not.  The solid and dashed edges indicate that the human does not intervene and intervenes, respectively.
\end{example}

We note that this turn-based game abstraction is able to capture multiple consecutive human moves 
as well as concurrent actions by the agents 
in the continuous domain with the assumption that human actions are faster than robot actions as in \cite{he2017reactive}. 
Prior work \cite{he2019automated} shows an automated process for constructing this abstraction and representing it in the Planning Domain Definition Language (PDDL) \cite{pddlbook}.




\subsection{Strategy and Payoff Function}

We assume both the human and robot are strategic agents, i.e., they choose actions in accordance with their own objectives. 
Informally, a strategy is a mapping that chooses a valid action to perform given the history of executions so far.  
A finite execution of the game is a finite sequence of states of $\mathcal{G}$. The set of all finite executions is denoted by $V^*$.

\begin{definition}[Strategy]
    
    A strategy for the robot is a function $\sigma: V^* \cdot V_s \to A_s$ that maps a finite execution ending in a robot state in $V_s$ to a robot action in $A_s$.
    Similarly, a strategy for the human is a function $\tau: V^* \cdot V_e \to A_e$ that maps a finite execution ending in a human state in $V_e$ to a human action in $A_e$.
    A strategy is said to be \textit{memoryless} if it only depends on the current state.  Otherwise, it is called a \textit{finite memory} strategy. We denote by $\Sigma^\G$ and $\Tau^\G$ the set of all strategies for the robot and the human, respectively, in $\G$.
    \label{def: sys_strategy}
\end{definition}



The realization of the strategies $\sigma$ and $\tau$ on abstraction $\mathcal{G}$ is called a play, which is a sequence of states
that correspond to the evolution of the world under the actions executed by the robot and human as per their strategies. 

\begin{definition}[Play]
    A \textit{play} (a.k.a \textit{run} or \textit{trajectory}) on $\mathcal{G}$ is an infinite sequence of states $r(\sigma, \tau) = v_0 v_1 v_2 \ldots \in V^{\omega}$ induced by strategies $\sigma$ and $\tau$ such that it is \textit{consistent} with the strategies, i.e.,  for all $i\geq 0$, $v_{i+1} = \delta(v_i,\sigma(v_0\ldots v_i))$ if $v_i \in V_s$, otherwise $v_{i+1} = \delta(v_i,\tau(v_0\ldots v_i))$.
    \label{def: two_player_game_play}
\end{definition}

For a play $r(\sigma, \tau) = v_0 v_1 \ldots$, we call the obtained sequence of observations $\rho(r(\sigma, \tau)) = L(v_0) L(v_1) \ldots$, where each $L(v_i) \in 2^{\Pi}$, the \textit{trace} of $r$.
We now define a payoff function that helps quantitatively reason over different plays in $\mathcal{G}$ induced by various strategies. Informally, the payoff function is the total energy cost that the robot incurs in a play.



 \begin{definition}[Payoff]
    Given robot and human strategies $\sigma$ and $\tau$, the \textit{payoff function} at state $v_0$ 
    is the total cost of the robot actions in the induced play $r(\sigma,\tau) = v_0 v_1 \ldots$, 
    i.e,
    \begin{equation}
        \Val^{v_0}(\sigma,\tau) = \sum_{i = 1}^{\infty} F(v_{i-1}, a_i, v_i),
    \end{equation}
    where 
    $a_i = \sigma(v_0 \ldots v_{i-1})$ if  $v_{i-1} \in V_s$, otherwise $a_i = \tau(v_0 \ldots v_{i-1})$.
    \label{def: payoff_function}
\end{definition}

\begin{example}
For abstraction $\G$ in Fig. \ref{fig: Abstraction illustration}, a memoryless strategy $\tau$ for the human is to always intervene ($a_{e1}$) while a finite-memory strategy $\tau$ is to not intervene ($a_{e2}$) if the strategy $\sigma$ for the robot is to build the arch near the human. 
The payoff $\Val$ associated with the strategy that the robot always picks action $a_{s2}$ and human picks action $a_{e2}$
is 5.
\label{eg: strategy_example}
\end{example}

\subsection{Manipulation Task}
\label{ssec: manipulation_task}

We consider manipulation tasks that can be achieved in finite time.  To express such tasks, we use LTLf \cite{vardi2013ltlf}, which is a language that combines Boolean connectives with temporal operators, allowing expression of complex tasks. 

\begin{definition}[LTLf Syntax \cite{vardi2013ltlf}] The syntax of LTLf formula $\phi$ is defined recursively as:
    \begin{equation}
        \phi :=  p \; | \neg \phi \;| \phi\; \wedge\; \phi \;| X\; \phi\; \;| \phi\, U \phi 
    \end{equation}
    where $p \in \Pi$ is an atomic proposition, ``$\neg$" (negation) and $``\wedge$" (and) are Boolean operators, and ``$X$" (next) and ``$U$" (until) are the temporal operators. 
    \label{def: ltlf_syntax}
\end{definition}

\noindent
The commonly used temporal operators ``$F$" (eventually) and ``$G$" (globally) are defined as: 
$F \phi := \top \; U\; \phi$ and $G \phi := \neg F \neg \phi$.
The LTLf semantics is as follows.
\begin{definition}[LTLf Semantics \cite{vardi2013ltlf}] LTLf formulas are defined over finite sequence of observations called a trace $\rho \in~(2^{\Pi})^{\ast}$. Let $|\rho|$ denote the length of the trace $\rho$, $\rho[i]$ be the $i^{th}$ observation in $\rho$, and $\rho,i\; \vDash\; \phi$ denote that the $i^{th}$ step of trace $\rho$ that satisfies the formula $\phi$. Then,
    \begin{itemize}
        \item $\rho,i\; \vDash\; \top$,
        \item $\rho,i\; \vDash\; p\;$ iff $p\; \in\; \rho[i]$,
        \item $\rho,i\; \vDash\; \neg \phi$ iff $\rho,\; i\; \nvDash\; \phi$,
        \item $\rho,i\; \vDash\; \phi_1\; \wedge\; \phi_2$ iff $\rho,\; i\; \vDash\; \phi_1$ and $\rho,\; i\; \vDash\; \phi_2$,
        \item $\rho,i\; \vDash\; X \phi$ iff $|\rho|\; >\; i + 1$ and $\rho,\; i+1\; \vDash\; \phi$,
        \item $\rho,i\; \vDash\; \phi_1\; U\; \phi_2$ iff $\exists\; j$ s.t. $i \leq j < |\rho|$, and $\rho,\; i\; \vDash\; \phi_2$ and $\forall k,\: i \leq k < j, \rho,\; k\; \vDash\; \phi_1$.
    \end{itemize}
    \label{def: ltl_semantics}
\end{definition}

\noindent
The set of finite traces that satisfy $\phi$ is called the language of $\phi$, i.e., $\lang(\phi) = \{\rho \in (2^\Pi)^{\ast}\; |\; \rho \vDash \phi\}$.  

We say that a play $r(\sigma,\tau)$ of $\G$ satisfies LTLf formula $\phi$ iff there exists a prefix of its trace $\rho(r(\sigma,\tau))$ that satisfies $\phi$, i.e., 
$r(\sigma,\tau) \models \phi \; \text{ iff }$
$$\; \exists \, pre(\rho(r(\sigma,\tau))) \; \text{ s.t. }  \; pre(\rho(r(\sigma,\tau))) \models \phi,$$ 
where $pre(\rho(r(\sigma,\tau)))$ is a prefix of trace $\rho(r(\sigma,\tau))$.  



\begin{example}
\label{example: arch formula}
For the arch-building example in Fig. \ref{fig: reg_illustration}, the task can be written as the LTLf formula
\begin{align*}
    \phi_\text{arch} =& F \left( p_\text{green, top} \wedge p_{\text{block, support}_{1}} \wedge p_{\text{block, support}_2} \right) \wedge \\ 
    & G \left( \neg(p_{\text{block, support}_1} \wedge p_{\text{block, support}_2}) \rightarrow \neg p_\text{green, top} \right).
\end{align*}
\end{example}


\subsection{Problem}



We are interested in generating robot strategies that encourage collaboration with humans.  At the same time, we require the robot to complete the task while never exceeding a given energy budget.  We consider the following problem.

\begin{problem}
    \label{problem}
    Given abstraction $\mathcal{G}$, LTLf task specification $\phi$, and a user-defined energy budget $\mathcal{B}$, 
    compute a strategy $\sigma$ for the robot that not only guarantees completion of task $\phi$ but also \emph{explores possible collaborations} with the human while keeping the total energy  $\Val^{v_0}(\sigma,\tau) \leq \mathcal{B}$ for all $\tau \in \Tau^{\G}$.
\end{problem}


\section{Regret-based Reactive Synthesis}
\label{sec: approach}

In this section, we introduce our solution to Problem~\ref{problem}, by building on previous work \cite{he2017reactive} and formulating a regret game.  We discuss the intuition for the regret formulation to produce cooperation-seeking behaviors for the robot.
Our proposed approach first converts the LTLf formula $\phi$ to a discrete structure that graphically represents the task,
and then composes it with with abstraction $\G$.  
On the resulting structure, we formulate a regret game with a proper definition for regret that incorporates the desired attribute of seeking cooperation while guaranteeing task completion.

\subsection{DFA Game}
Every LTLf formula $\phi$ can be converted to a \emph{Deterministic Finite Automaton} (DFA) that captures all the possible ways in which one can satisfy $\phi$ \cite{vardi2013ltlf}. 
Given task $\phi$, we construct DFA $\mathcal{A}_\phi = (Z, z_0, \Sigma, \delta, Z_f)$, where $Z$ is a finite set of states,  $z_0$ is the initial state, $\Sigma = 2^\Pi$ is the alphabet, 
$\delta: Z \times \Sigma \rightarrow Z$ is the deterministic transition function, and $Z_f \subseteq Z$ is the set of accepting states.
A run of $\mathcal{A}_\phi$ on a trace $\rho = \rho[1] \rho[2] ...\rho[n]$, where $\rho[i] \in 2^{\Pi}$, is a sequence of DFA states $z_0 z_1 \ldots z_n$, where $z_{i+1} = \delta(z_i, \rho[i])$. 
If $z_n \in Z_f$, then trace $\rho$ satisfies $\phi$, i.e., $\rho \models \phi$.


Given a DFA $\mathcal{A}_\phi$ and the manipulation domain abstraction $\G$, we compose the two graphs to construct the DFA Game $\mathcal{P}$ that captures all the possible ways in which task $\phi$ can be accomplished on $\mathcal{G}$. The states in $\mathcal{P}$ represent the current configuration of the physical world as well as how much of the task $\phi$ has been accomplished so far. 

\begin{definition}[DFA Game]
A \textit{DFA Game} is a tuple $\PA = \G \times \mathcal{A}_\phi = (S, S_f, s_0, A_s, A_e, F_\PA, \delta_{\PA})$ where $A_s$ and $A_e$ are as in Def. \ref{def: two_player_game}, and
    \begin{itemize}
        \item $S = V \times Z$ is a finite set of states, and $S_s = V_s \times Z$ and $S_e = V_e \times Z$ are the set of robot and human states, respectively,
        \item $s_0 = (v_0, z_0)$ is the initial state,
        \item $S_f = \{(v, z)\, |\, z \in Z_f\}$ is the set of final or accepting states,
        \item $\delta_{\PA} : S \times (A_s \cup A_e) \rightarrow S$ is the deterministic transition function such that a transition from $s = (v, z)$ to $s'=~(v', z')$ under $a \in (A_s \cup A_e)$ exists if $v'= \delta(v, a)$ and $z' = \delta(z, L(v))$ and $z \not \in Z_f$.  If  $z' \in Z_f$, then $\delta_\PA((v,z),a) = (v,z)$ for all $a \in (A_s \cup A_e)$,
        \item $F_\PA: S \times (A_s \cup A_e) \times S \rightarrow \mathbb{R}_{\geq 0}$ is the cost function such that $F_\PA((v,z),a,(v',z')) = F(v,a,v')$.
    \end{itemize}
\label{def: product_automaton}
\end{definition}

A run on $\PA$ is a valid sequence of states obtained with $\delta_\PA$. Thus, its projection onto DFA $\mathcal{A}_\phi$ is a valid run of $\mathcal{A}_\phi$.
Thus a run that ends up in an accepting state in $S_f$ corresponds to also an accepting run in $\mathcal{A}_\phi$ and satisfies $\phi$.  
Therefore, the problem reduces to finding a robot strategy $\sigma \in \Sigma^\PA$ on $\PA$ that guarantees that the robot can enforce a visit to the accepting set $S_f$ under all possible human moves \cite{he2019automated}.


\begin{figure}[t]
    \centering
    \includegraphics[scale=0.25]{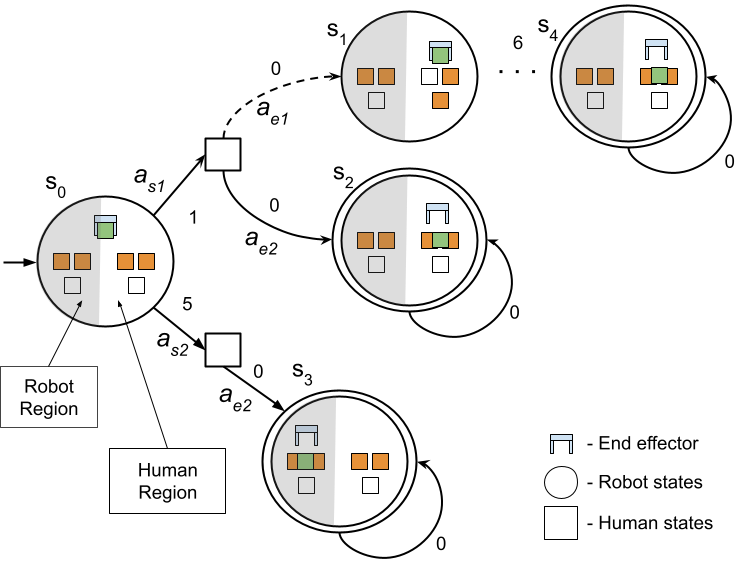}
    \caption{DFA Game $\PA$ for the abstraction $\G$ in Fig. \ref{fig: Abstraction illustration} and the arch-building task. 
    The double edged states denote accepting states. 
    When operating close to the human, the robot spends either 1 (cooperative human) or 7 (adversarial human) units of total energy. Otherwise, it spends 5 units of total energy.}
    \label{fig: dfa game illustration}
    \vspace{-5mm}
\end{figure}

\begin{example}
Fig. \ref{fig: dfa game illustration} illustrates a DFA game for the abstraction $\G$ in Fig. \ref{fig: Abstraction illustration} with the arch-building task.
Say, the energy budget for the robot 
 $\mathcal{B} = 7$. A strategy for the robot that guarantees task completion within the budget is to take action $a_{s2}$ from $s_0$, which requires 5 units of energy, irrespective of the human's action. This strategy is obtained if human is assumed to be adversarial.
\end{example}

\noindent
Below, we show how to formulate $\PA$ as a regret game to relax the adversarial assumption.

\subsection{Reactive Synthesis with no Regret}
\label{ssec: regret_synthesis}
An adversarial assumption is often a conservative abstraction of the reality because humans tend to be cooperative.
Thus, we relax this assumption and also account for the fact that the human can have their own objective. Rather than picking an action which is optimal for all possible human actions, we pick an action that is ``good enough" for all human actions. 
We propose regret to be a viable solution concept  
that quantifies how good an action is compared to the best possible action associated with the best outcome, i.e., the run with the least payoff in $\PA$.

To build the intuition, consider the DFA game in Fig.~\ref{fig: dfa game illustration}. The robot has two choices from the initial state $s_0$: to build the arch within human's reach (action $a_{s_1}$) or away from the human (action $a_{s2}$). The best action for the robot if the human decides to intervene is action $a_{s2}$ and spend 5 units of total energy. If robot takes action $a_{s1}$, it spends  
1 unit of energy
for a cooperative human and 7 units of energy for an intervening human.  Then, we say the ``regret'' associated with robot action $a_{s1}$ is 2 if the human intervenes because the robot could have achieved energy 5 with $a_{s2}$. If the human does not intervene, then regret of $a_{s1}$ is 0. Similarly, if the robot decides to take action $a_{s2}$ and the human decides to intervene, regret is 0, otherwise it is 4. Table \ref{tab: table_reg_vals} summarizes the regret values associated with all strategies. A regret-minimizing strategy for the robot is then to build the arch near the human (action $a_{s1}$) as it can guarantee a minimum regret of 2 under all human actions.
Intuitively, a regret-minimizing action for this task is to give the human a chance to be cooperative as it can still achieve the task with either actions while stay within the energy budget.
We now formally define \textit{regret}. 




\subsubsection{Regret}
\label{sssec: regret}

\begin{table}[t!]
    \caption{The regret associated with each robot and human action from \eqref{eq: reg_def} for DFA-Game $\PA$ in Fig. \ref{fig: dfa game illustration}}
    \centering
    \begin{tabular}{|c||*{2}{c|}}\hline
    \backslashbox{Human ($\tau$)}{Robot ($\sigma$)}
    & \makebox[2 cm]{near ($a_{s1}$)}&\makebox[2 cm]{away ($a_{s2}$)}\\\hline
    Intervene ($a_{e1}$)  & 7 - 5 = 2  & 5 - 5 = 0 \\\hline
    No Intervene ($a_{e2}$)  & 1 - 1 = 0 & 5 - 1 = 4\\\hline
    \end{tabular}
\label{tab: table_reg_vals}
\vspace{-3mm}
\end{table}




We define
regret 
as the \emph{difference} in the outcome (total payoff) associated
with the current action and the best possible outcome, i.e., if the human is purely cooperative. We refer to the best possible outcome as \textit{best-response}.
Thus, we use best-responses as yardsticks to compare the quality of each robot action.

\begin{definition}[Task-Aware Regret]
    \label{def: reg_def}
    Given strategies $\sigma$ and $\tau$ on DFA game $\PA$ for the robot and human, respectively, 
    \textit{task-aware regret} at state $s$ is defined as the difference in the payoff $\Val^{s}(\sigma, \tau)$ 
    and the best possible payoff $\min_{\sigma'} \Val^{s}(\sigma', \tau)$ 
    under the same human strategy $\tau$, i.e., 
    \begin{equation}
        \reg^s(\sigma, \tau) = \Val^{s}(\sigma, \tau) - \min_{\sigma'}\, \Val^{s}(\sigma', \tau).
        \label{eq: reg_def}
    \end{equation}
\end{definition}
\noindent
Here $\sigma'$ is any alternate strategy for the robot other than $\sigma \,$, i.e., $\sigma' \in \Sigma^\PA \setminus \{\sigma\}$. We say that $\reg^s(\sigma) = + \infty$ if $\sigma' \in \emptyset$.

We use best-responses ($ \min_{\sigma'} \Val^{s}(\sigma', \tau)$) to compare the quality of each robot action for a fixed  $\tau$ of the human. As the regret behaviors change according to the task at hand, we call them task-aware regret. 
Hence, our goal is find a strategy $\sigma^{\ast}$ that behaves ``not far" from an optimal response to the strategy of the human $\tau$ when $\tau$ is fixed. Thus, 
\begin{align}
    \label{eq:reg-min-strategy}
    \sigma^{\ast} = \arg \min_{\sigma} ( \max_\tau \reg^{\sigma, \tau}(s_0)).
\end{align}


In Table \ref{tab: table_reg_vals}, the combination of strategies $(a_{s1}, a_{e2})$ and $(a_{s2}, a_{e1})$ represent the extreme scenarios. A regret minimizing strategy $\sigma^*$ is to pick a path with lower regret value - the worst-case payoff for current strategy $\sigma$ is similar to the payoff of the best alternative $\sigma'$. Thus, we pick action $a_{s1}$ as it has lower regret value. Note that the play induced by action $a_{s1}$ also has a path, in which the human can cooperative and help the robot spend less amount of energy.
Thus, the robot seeks for strategies that have similar best alternative payoffs from the current state. If the human is cooperative then the robot has more energy to spare and hence gives the human more opportunities to collaborate before finishing the task. The minimum budget $\mathcal{B}$ required to synthesize a regret-minimizing strategy is the total payoff associated with the winning strategy for the robot to complete the task assuming human to be purely adversarial.



We note that a key difference in our regret formulation from the ones used in the machine learning community is that we look at the set of all strategies, including the finite-memory ones for both the human and the robot.
 In the learning community, they use regret to compare the performance of their algorithms against a subset of strategies or a fixed adversary. 
For us, finite memory strategies are useful in keeping
track of past human actions and thus
allowing the robot to reason over the human's intention. 
\vspace{-2 mm}
\subsection{Computing Regret-Minimizing Strategies}
\label{ssec: regret_str_synthesis}

To compute for $\sigma^*$ in \eqref{eq:reg-min-strategy}, we use the results in \cite{filiot2010iterated}.  
The method is summarized in Algorithm 1, which  consists of two main steps.  
First, we compute all the plays
induced by all robot strategies $\sigma \in \Sigma^\PA$ for a fixed human strategy $\tau \in \Tau^\PA$ by \textit{unfolding} $\PA$ until the payoff of $\mathcal{B}$ is reached. This process gives us a tree-like structure. 
We then augment each node with the value of the total-energy spent by the robot ($\Val$) to reach that node. We call this graph the \textit{Graph of Utility} - $\G^{u}$ (Line~1). It
captures all the possible plays for a fixed human strategy $\tau$.
Note that paths of $\G^{u}$ are realizations of finite memory strategies. Furthermore, if an accepting state in $\PA$ is reached during unfolding, it is a leaf node in $\G^{u}$.

For each edge in every path in $\G^{u}$, we compute the best-alternate response ($BA$) by finding the lowest payoff that corresponds to $\min_{\tau} \min_{\sigma'}\, \Val^{s}(\sigma', \tau)$
(Line~4). We augment each node in $\G^{u}$ with the best-alternate response value and construct $\G^{br}$ (Line~6). 
For node $s \in \G^{br}$, the difference between its payoff and its best-alternate response
is the regret value $\reg^s$ at $s$. We repeat this process for all nodes in $\G^{br}$. Then, we run a value iteration based method to back propagate the $\reg^s$ values from the leaf nodes in $\G^{br}$ until we reach a fixed point. 

 The number of states of $\G^{br}$ (denoted by $|\G^{br}|$) is polynomial in the size of $\PA$. The algorithm is polynomial in the size of $\PA$ times $\B$.
 The memory of strategies is directly related to the depth of the tree and the number of alternate edges along that path. 

\begin{algorithm}[t]
    \caption{Compute regret minimizing strategies
    }
    \SetKwInOut{Input}{Input}\SetKwInOut{Output}{Output}
    \SetKwData{Edges}{edges}
    \SetKwData{This}{this}\SetKwData{Up}{up}
    \SetKwFunction{GraphOfUtility}{GraphOfUtility}
    \SetKwFunction{ComputeBA}{ComputeBA}
    \SetKwFunction{GraphOfBestResponse}{GraphOfBestResponse}
    \SetKwFunction{ValueIteration}{ValueIteration}

    \Input{DFA Game $\PA$ and energy budget $\mathcal{B}$}
    \Output{Regret minimizing strategy $\sigma^{\ast}$}
    
    $\G^{u} \gets$ \GraphOfUtility($\PA$, $\B$) \\
    
    $BA[e] \gets$ $\infty$ for all \Edges in $\G^{u}$\\
    \For{all \Edges $\in \G^{u}$}{

        $BA[e] \gets$  \ComputeBA(\Edges)\\ 
    }
    $\G^{br} \gets$ \GraphOfBestResponse($\G^{u}$, $BA$)\\
    $\sigma^{\ast} \gets$  \ValueIteration($\G^{br}$)\\ 
    \KwRet{$\sigma^{\ast}$}
    
        
        
    \label{algo: compute_regret}
\end{algorithm}
\vspace{-1mm}
\section{Experiments}
\label{sec: case_studies}
\vspace{-1mm}

\begin{figure*}[t]
    \centering
    \begin{subfigure}[t]{0.135\textwidth}
        \centering
        \includegraphics[width=0.99\linewidth]{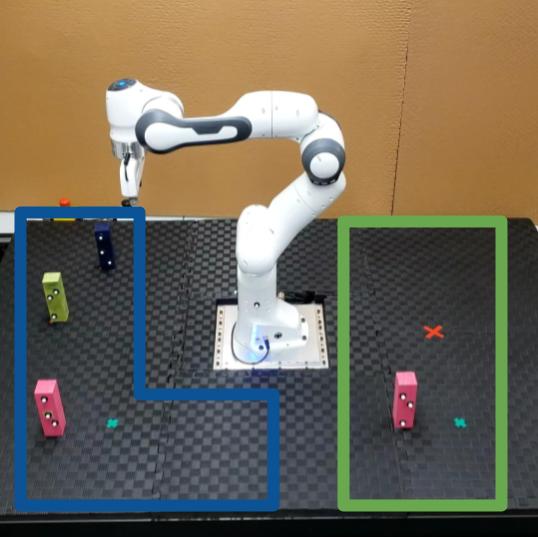}
        \caption{}
    \label{fig: arch_regret_a}
    \end{subfigure}%
    ~ 
    \begin{subfigure}[t]{0.135\textwidth}
        \centering
        \includegraphics[width=0.99\linewidth]{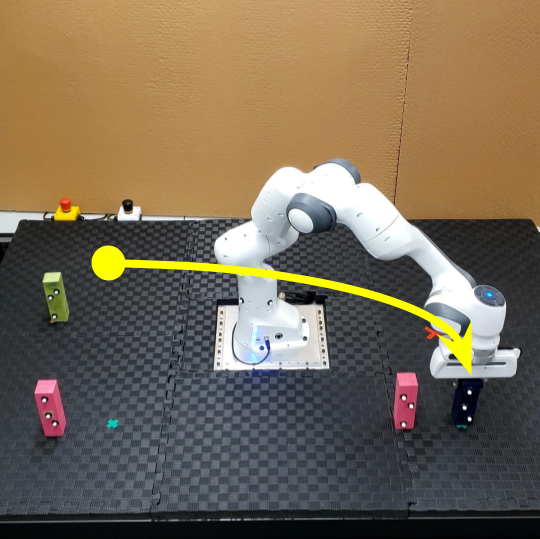}
        \caption{}
    \label{fig: arch_regret_b}
    \end{subfigure}%
    ~
    \begin{subfigure}[t]{0.135\textwidth}
        \centering
        \includegraphics[width=0.99\linewidth]{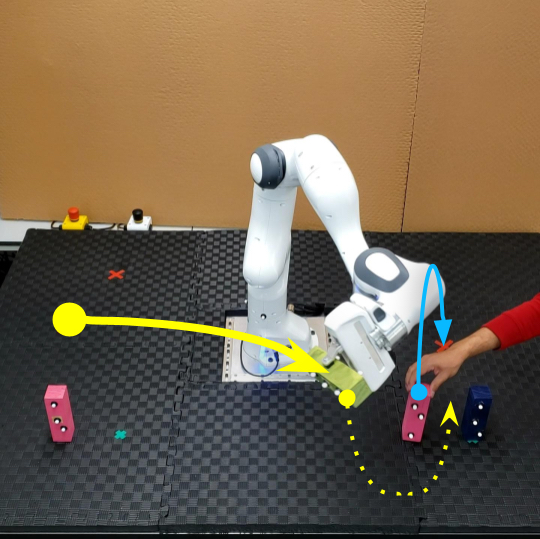}
        \caption{}
    \label{fig: arch_regret_c}
    \end{subfigure}%
    ~ 
    \begin{subfigure}[t]{0.135\textwidth}
        \centering
        \includegraphics[width=0.99\linewidth]{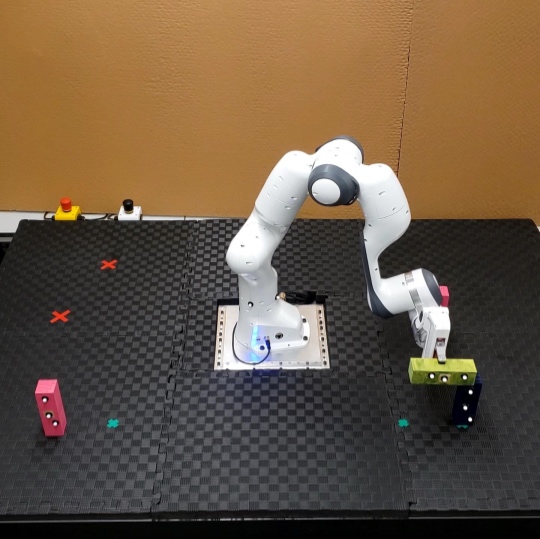}
        \caption{}
    \label{fig: arch_regret_d}
    \end{subfigure}%
    ~ 
    \begin{subfigure}[t]{0.135\textwidth}
        \centering
        \includegraphics[width=0.99\linewidth]{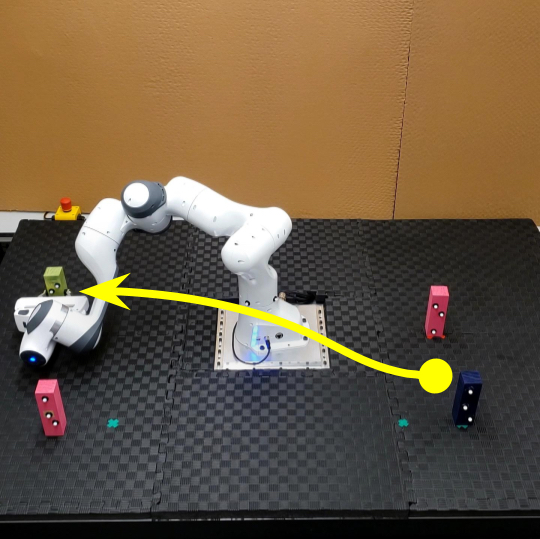}
        \caption{}
    \label{fig: arch_regret_e}
    \end{subfigure}%
    ~ 
    \begin{subfigure}[t]{0.135\textwidth}
        \centering
        \includegraphics[width=0.99\linewidth]{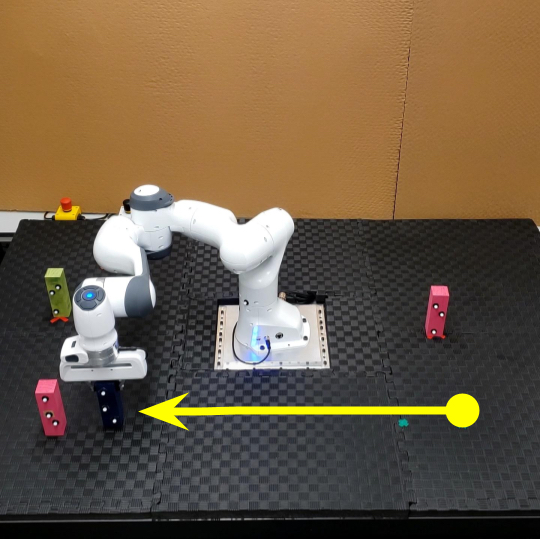}
        \caption{}
    \label{fig: arch_regret_f}
    \end{subfigure}%
    ~ 
    \begin{subfigure}[t]{0.135\textwidth}
        \centering
        \includegraphics[width=0.99\linewidth]{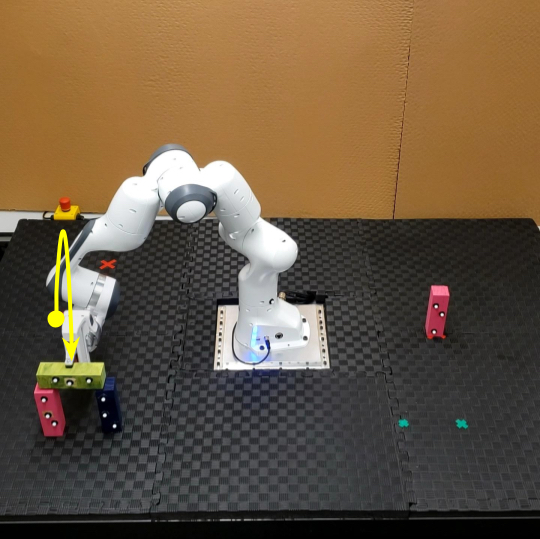}
        \caption{}
    \label{fig: arch_regret_g}
    \end{subfigure}%
    
\caption{Arch construction with one human intervention. 
Human and robot regions are indicated in green and blue respectively in (a).
Yellow and blue arrow represent robot and human actions. Video: \small{youtu.be/ABZb1g36Kv4}
}
\label{fig: arch regret}
\vspace{-4mm}
\end{figure*}

\begin{figure}[t]
    \centering
    \begin{subfigure}[t]{0.14\textwidth}
        \centering
        \includegraphics[width=0.97\linewidth]{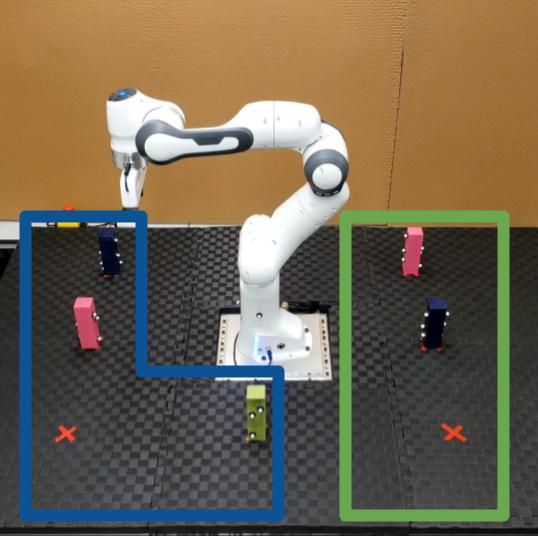}
        \caption{}
    \label{fig: line_regret_a}
    \end{subfigure}%
    ~ 
    \begin{subfigure}[t]{0.14\textwidth}
        \centering
        \includegraphics[width=0.97\linewidth]{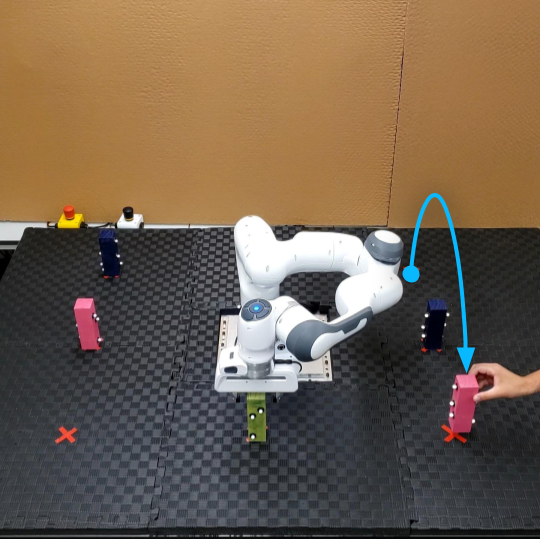}
        \caption{}
    \label{fig: line_regret_b}
    \end{subfigure}%
    ~
    \begin{subfigure}[t]{0.14\textwidth}
        \centering
        \includegraphics[width=0.97\linewidth]{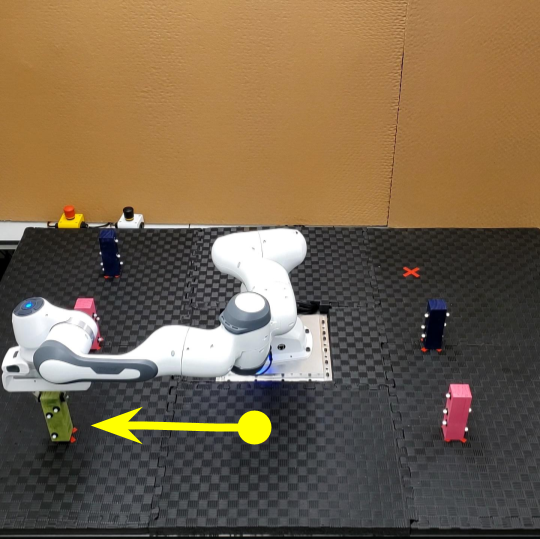}
        \caption{}
    \label{fig: line_regret_c}
    \end{subfigure}%
    
    \begin{subfigure}[t]{0.14\textwidth}
        \centering
        \includegraphics[width=0.97\linewidth]{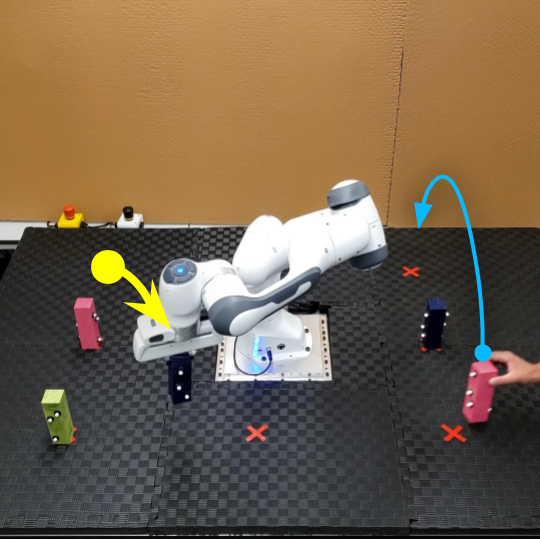}
        \caption{}
    \label{fig: line_regret_d}
    \end{subfigure}%
    ~ 
    \begin{subfigure}[t]{0.14\textwidth}
        \centering
        \includegraphics[width=0.97\linewidth]{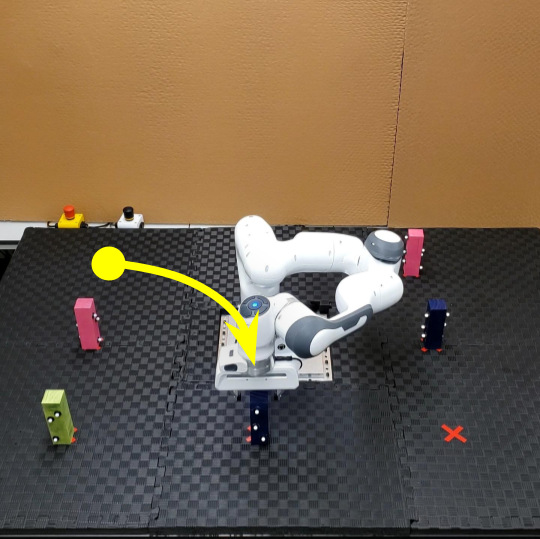}
        \caption{}
    \label{fig: line_regret_e}
    \end{subfigure}%
    ~ 
    \begin{subfigure}[t]{0.14\textwidth}
        \centering
        \includegraphics[width=0.97\linewidth]{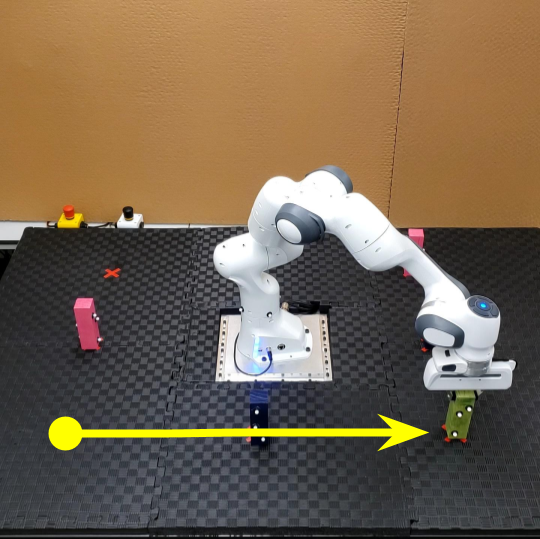}
        \caption{}
    \label{fig: line_regret_f}
    \end{subfigure}%
    
    \vspace{-2mm}
\caption{Straight line alignment with two human interventions. Robot (blue) and human (green) regions are shown in (a).
}
\label{fig: line regret}
\vspace{-5mm}
\end{figure}

We illustrate the efficacy of our framework on two different case studies analogous to the ones in \cite{he2017reactive, he2019automated}.
In each scenario the robot has two regions to complete the task. One is near the human, who may intervene or collaborate with the robot. The other region is far from the human, 
where they cannot reach. 
The robot spends 3 unit of energy per action to operate in this region whereas 1 unit of energy near the human and
was given a fixed budget $\B$ to complete the task. 

We constructed an abstraction for each scenario as in \cite{he2017reactive,he2019automated} 
(see \cite{muvvala2021thesis} for details). 
Then, we used our regret-based synthesis framework to generate a strategy and 
compared the emergent behaviors against the behaviors seen when using the framework in \cite{he2017reactive}, which assumes the human is purely adversarial. 
Fig. \ref{fig: arch regret} and \ref{fig: line regret} show example behaviors, and Table \ref{tab: tabular_summary_case_studies} outlines the computational costs.
Videos of all the case-studies are available to view in \cite{videos}.


The implementation of our framework is an end-to-end software tool that takes in the manipulation domain abstraction 
and task 
LTLf formula and generates a regret-minimizing strategy. The tool is in Python and is available on GitHub \cite{repolink}.  
For the experiments, the strategy was implemented on the Franka Emika Panda robotic manipulator.

\textit{\textbf{Arch Construction:} \quad}
In the first scenario, the task 
is to build an arch with the green box on top (LTLf formula shown in Example \ref{example: arch formula}).
The total budget was $\mathcal{B} = 10$, and the total energy needed to finish the task away from the human is 4 units. 
Fig. \ref{fig: line regret} shows an execution of this task with one human intervention. 
The robot initially starts to build the arch near the human as shown in Fig. \ref{fig: arch_regret_a}-\ref{fig: arch_regret_c}. 
But, the human intervenes adversarially by moving the support (pink object) away.  Then, the robot becomes conservative and builds the arch in the other region, spending 6 units of energy. A purely adversarial behavior for the robot is to build the arch away from the human, thus spending 4 units of total energy. 
Note that the robot ends up spending more energy under the regret-minimizing strategy, but it still stays within its energy budget while seeking collaboration.  The general trend for such strategies is to be optimistic and seek collaboration with the human until the human disrupts the cooperation, motivating the robot to become pessimistic to ensure completing the task within the given energy budget.  

\textit{\textbf{Straight Line Alignment}: \quad}
The second task for the robot is put three objects in a line such that pink block is in the top location, the blue block is in the middle, and the green block is at the bottom.
Fig. \ref{fig: line regret} shows an execution of this task. 
Note that the robot needs to execute fewer actions to accomplish the task near the human. An adversarial strategy for the robot is to rearrange the blocks placed away from the human irrespective of the human's action. However, a regret minimizing strategy is to consider the possibility that the human could be cooperative. For this scenario $\mathcal{B} = 20$ and the total-energy to finish the task away from the human is 12, and 2 if the human is cooperative. Initially, the human intervenes, adversarially, thus the robot operates away from the human (Fig. \ref{fig: line_regret_a}-\ref{fig: line_regret_c}). The human then intervenes again and opens up another opportunity for cooperation (Fig. \ref{fig: line_regret_d}-\ref{fig: line_regret_f}). The robot takes this opportunity and completes the task.

\textit{\textbf{Computational Cost}: \quad}
As the total energy required to accomplish the task without any human cooperation increases, we see that the robot provides more opportunities for the human to be cooperative as long as it can still guarantee to complete the task while staying below the energy budget $\mathcal{B}$.
However, it comes at a cost.  As shown in Table \ref{tab: tabular_summary_case_studies}, as $\mathcal{B}$ increases, more memory is required to generate the strategy.  Therefore, it becomes computationally more expensive.
\begin{table}[h!]
    \vspace{-1mm}
    \caption{Total number of states in various abstractions, energy budget $\mathcal{B}$, and average runtime and memory usage.}
    \label{tab: tabular_summary_case_studies}
    \centering
    \begin{tabular}{|p{0.5 cm}|| c | c | r | r | c | p{0.8cm} |} 
         \hline
         \small{Case study} & $|S_\PA|$ & $\mathcal{B}$ & $|{\G^{u}}|\;\;\,$ & $|{\G^{br}}|\;\;\;\,$ & Time (s) & Memory (GB)\\ 
         \hline
         \multirow{5}{*}{Arch} & \multirow{5}{*}{48,843} & 10 & 537,274 & 2,152,851 & 405 &\;\;6.12 \\
         & & 12 & 634,960 & 3,625,717 & 515 & \;\;8.26\\
         & & 14 & 732,721 & 5,477,123 & 622 & 12.10 \\
         & & 16 & 830,417 & 7,755,377 & 756 & 16.80\\
         & & 18 & 928,113 & 10,469,395 & 898 & 21.88 \\
         \hline
         \multirow{5}{*}{Line} &  \multirow{5}{*}{19,254} & 15 & 308,065  & 2,306,785  & 174 &\;\;5.03 \\
         &  & 17 & 346,573 & 3,263,221 & 221 &\;\;7.01 \\
         &  & 19 & 385,081 & 4,368,121 & 273 &\;\;8.89 \\
         &  & 21 & 423,589 & 5,621,485 & 328 & 11.72 \\
         &  & 23 & 462,097 & 7,023,313 & 371 & 13.91 \\
         \hline
    \end{tabular}
    \vspace{-3mm}
\end{table}

\section{CONCLUSION}
\label{sec: conclusion}

We presented a different formulation for synthesizing strategies for a robot operating in presence of a human. We use regret to relax the adversarial assumption on human and allow the robot to seek collaboration. We find the emergent behavior for the robot to be more intuitive and human-like. For future work, we plan to extend this framework to more complex scenarios with multiple agents as well as 
improving the computational cost of algorithm for efficient synthesis.


\newpage

\bibliographystyle{IEEEtran}
\bibliography{references}

\end{document}